\documentclass[pdflatex,sn-mathphys-num]{sn-jnl}
\usepackage[utf8]{inputenc}

\usepackage{graphicx}%
\usepackage{multirow}%
\usepackage{amsmath,amssymb,amsfonts}%
\usepackage{amsthm}%
\usepackage{mathrsfs}%
\usepackage[title]{appendix}%
\usepackage{xcolor}%
\usepackage{textcomp}%
\usepackage{manyfoot}%
\usepackage{booktabs}%
\usepackage{algorithm}%
\usepackage{algorithmicx}%
\usepackage{algpseudocode}%
\usepackage{listings}%
\usepackage{pifont} 

\usepackage[T1]{fontenc}
\usepackage{xcolor}
\definecolor{gainGreen}{HTML}{228B22}
\definecolor{lossRed}{HTML}{C00000}

\usepackage{booktabs}   
\usepackage{array}      

\usepackage{subcaption}            
\usepackage{hyperref}   

\usepackage{lineno}
\modulolinenumbers[1]        

\theoremstyle{thmstyleone}%
\theoremstyle{thmstyletwo}%

\theoremstyle{thmstylethree}%

\begin{document}


\title{Hierarchical Graph Networks for Accurate Weather Forecasting via Lightweight Training}

\author*[1]{\fnm{Thomas} \sur{Bailie}}\email{thomas.bailie@auckland.ac.nz}

\author[2,3]{\fnm{S. Karthik} \sur{Mukkavilli}}\email{drkarthik@kaist.ac.kr}

\author[4]{\fnm{Varvara} \sur{Vetrova}}\email{varvara.vetrova@acu.edu.au}

\author[1]{\fnm{Yun Sing} \sur{Koh}}\email{y.koh@auckland.ac.nz}

\affil*[1]{\orgdiv{School of Computer Science}, \orgname{The University of Auckland}, \orgaddress{ \city{Auckland}, \country{New Zealand}}}

\affil[2]{\orgdiv{GGGS}, \orgname{Korea Advanced Institute of Science \& Technology}, \city{Daejeon}, \country{South Korea}}

\affil[3]{\orgname{Mercuria Energy Group}, \city{Geneva}, \country{Switzerland}}

\affil[4]{\orgname{Australian Catholic University}, \city{Sydney}, \country{Australia}}







\abstract{

        Climate events arise from intricate, multivariate dynamics governed by global-scale drivers, profoundly impacting food, energy, and infrastructure. Yet, accurate weather prediction remains elusive due to physical processes unfolding across diverse spatio-temporal scales, which fixed-resolution methods cannot capture. Hierarchical Graph Neural Networks (HGNNs) offer a multiscale representation, but nonlinear downward mappings often erase global trends, weakening the integration of physics into forecasts. We introduce HiFlowCast and its ensemble variant HiAntFlow, HGNNs that embed physics within a multiscale prediction framework. Two innovations underpin their design: a Latent-Memory-Retention mechanism that preserves global trends during downward traversal, and a Latent-to-Physics branch that integrates PDE solution fields across diverse scales. Our Flow models cut errors by over 5\% at 13-day lead times and by 5–8\% under 1st and 99th quantile extremes, improving reliability for rare events. Leveraging pretrained model weights, they converge within a single epoch, reducing training cost and their carbon footprint. Such efficiency is vital as the growing scale of machine learning challenges sustainability and limits research accessibility. Code and model weights are in the supplementary materials.
}


\maketitle
    
\section{Introduction}\label{sec1}

Weather and climate variability govern critical dimensions of human life. Rainfall and humidity regulate crop yields \cite{challinor2014climate, ray2015climate}, directly shaping the global food supply. Wind patterns influence the reliability of renewable energy production \cite{emeis2018wind}, while extreme events such as cyclones can inflict widespread damage, threatening both infrastructure and livelihoods. Anticipating these outcomes is therefore central to food and energy security. However, accurate medium-range weather forecasting, on lead times of days to weeks, remains scientifically challenging due to the intricate, multiscale nature of the climate system.

Multivariate weather forecasting \cite{reichstein2019deep,bi2023accurate} seeks to extend projections far into the future, until forecasts violate governing physics principles. In data-driven settings, this infeasibility manifests once the error between ground truth and prediction surpasses a critical threshold. Long lead-time forecasts provide the scientific community with an accessible tool, for instance, to explore ocean dynamics \cite{patil2023enso}. Yet, forecasting remains challenging due to the intrinsic complexity of the climate system, which arises from interactions spanning multiple spatio-temporal scales. For example, El Niño–La Niña cycles \cite{McPhaden2006,Meehl2001} drive large-scale ocean circulation while simultaneously shaping local-scale phenomena. Consequently, even short lead-time forecasts are susceptible to high error, which compounds rapidly at longer lead times. A major source of this error is the partial modeling of these intricate climate dynamics, preventing the capture of interactions between multiscale processes. 

In weather forecasting, Hierarchical Graph Neural Networks (HGNNs) \cite{oskarsson2024probabilistic} represent multiscale phenomena as interacting processes, enabling global dynamics to inform local projections. By explicitly modeling the interplay between global and regional processes, HGNNs yield a higher-fidelity computational representation of the climate system than single-resolution or multi-mesh methods. Yet, current HGNNs fall short of their potential. During downward traversal, excessive compression erases critical multiscale physics \cite{topping2022understanding}, leaving only a partial account of the drivers of local dynamics. This loss constrains the high-correlation lead times achievable by HGNNs. While residual connections can partly preserve global trends, they are restrictive in their degrees of freedom. Specifically, they exploit only downward-directed features rather than jointly integrating information from both upward and downward traversals. Consequently, local phenomena are only partially contextualized, weakening the coherence between global and local processes.

\begin{figure*}[ht]
    \centering
    \includegraphics[width=0.89\linewidth]{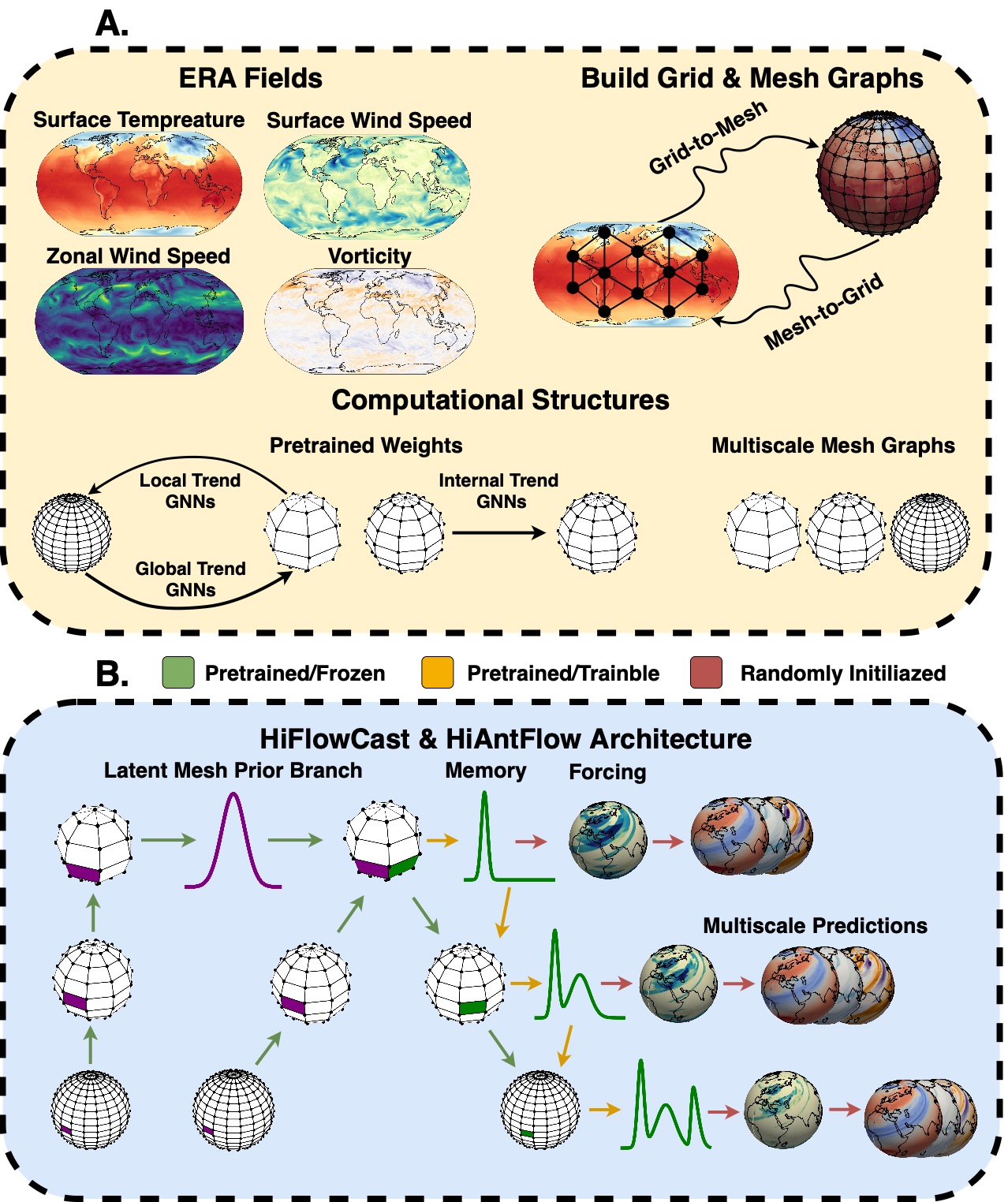}
    \caption{A. Latitude–longitude grids are projected onto a periodic mesh graph, preserving global continuity. HiFlowCast and HiAntFlow load pretrained hierarchical weights. B. Both models reuse pretrained components, freezing some to stabilize training. A memory buffer retains global trends, while a physics branch injects ground-truth signals at each level, enabling multiscale forecasts.}
    \label{fig:arch_fig}
\end{figure*}

The combination of high spatial resolution, extended evaluation intervals, and expansive parameter spaces also renders training data-driven weather forecasting models prohibitive via limited computational budgets \cite{rasp2020weatherbench}. Recent approaches illustrate the scale of these costs: ClimaX \cite{nguyen2023climax} requires pretraining on heterogeneous climate data with up to 80 NVIDIA V100 GPUs, while Pangu-Weather \cite{bi2023pangu} demands nearly 3000 V100 GPU-days. Such computational expenses create significant barriers for many research groups, underscoring the need for more efficient training strategies. Beyond these considerations, training requirements impose a substantial environmental burden. Large-scale neural networks emit the equivalent of several tonnes of CO\textsubscript{2} during training \cite{patterson2021carbon,luccioni2023estimating}, with life-cycle analyses further revealing that deployment and embodied emissions amplify this footprint \cite{faiz2023llmcarbon}. By reducing model size and trainable parameters, and adopting efficient training strategies, our approach aligns with the 4M best practices of \citet{patterson2022carbon} that can lower training energy by up to 100× and emissions by up to 1000×.

{\bf Our Contributions:} We introduce HiFlowCast and HiAntFlow, HGNNs designed to strengthen the influence of global-scale physics on climate projections. The models comprise two key innovations, see Figure \ref{fig:arch_fig} for a visual illustration. First, a Latent-Memory-Retention mechanism jointly models upward and downward directional features, preserving learnt global trends while maintaining local context. Second, a Latent-to-Physics branch blends PDE solution fields with the memory buffer features at each hierarchical level, partially bypassing the need for hierarchical traversal by forming a scale-constrained prediction. The branch, as a consequence, creates output modules specializing in mixing physics at multiple resolutions. Instead of embedding PDE solution fields solely at their native resolution, multiscale embedding enforces constraints that reflect the climate system's influence on processes across various resolutions. The innovations from these two branches enable HiFlowCast and HiAntFlow to unroll projections that integrate multiscale physics with global trends preserved in the memory buffer.

We address the considerable computational burden of training global weather forecasting models. Drawing on the universality of transformer architectures \cite{yun2019transformers} and transfer learning \cite{zhuang2020comprehensive}, we show that graph-based models pretrained on global climate datasets can be adapted to new architectures with little to no parameter tuning. Figure \ref{fig:arch_fig} highlights components that exploit universality (green), those that rely on transfer learning (yellow), and those that are randomly initialized (red). Using this strategy, we train HiFlowCast and HiAntFlow in a single epoch on a single A100 GPU, in place of $100$ to $200$ epochs it would otherwise take \cite{oskarsson2024probabilistic}, reducing training time by roughly two orders of magnitude. We provide an asymptotic analysis that bounds the additional evaluation cost by the size of interaction networks at the lowest hierarchical level, rather than by hierarchical depth.

Through empirical evaluation, we benchmark HiFlowCast and HiAntFlow against recent GNN-based weather forecasting models \cite{lam2023graphcast, oskarsson2024probabilistic}. Both Flow models frequently retain predictive skill up to ten days ahead, a horizon at which competing approaches degrade. Over this extended lead time, they improve relative performance by an average value of approximately $5$\% across MAE, CRPS, and RMSE when these metrics are averaged over all grid cells and normalized variables. In particular, substantial gains are due to improvements in variables tightly coupled to the solar radiation embedding: atmospheric and surface temperatures, specific humidity, and geopotential. We further assess the Flow model's performance over extremes, showing consistent gains in the $99$th quantile, with improvements of about $8$\% at a $13$-day lead time.

\section{Results}


\captionsetup[subfigure]{labelformat=parens,labelsep=space,skip=2pt}

We conduct a rigorous evaluation of HiFlowCast and HiAntFlow on global forecasting tasks to assess their reliability as global weather forecasting models. In particular, our analysis examines the fidelity of projected outcomes at a long lead time. In the ideal case, a skillful model will create strategies for approaching climate scenarios even at long lead times. We evaluate on the ERA5 reanalysis dataset \cite{hersbach2020era5}, which assimilates diverse observational records into a numerical weather prediction model to produce spatially and temporally consistent gridded fields. The dataset spans from 1959 to 2023. We train from 1959 to 2010 and evaluate generalizability in two non-overlapping holdout years: 2021 to 2022 for validation and 2022 to 2023 for testing. Consistent with prior work \cite{oskarsson2024probabilistic}, our version of ERA5 occupies a $5.625^\circ$ latitude–longitude grid. We use various metrics such as RMSE, MAE, and CRPS to conduct our evaluation, thereby studying the Flow models using long lead time analysis. Furthermore, models must remain reliable over extreme events. We evaluate all models on single-sample high and low-intensity extreme weather events over temperature and wind speed variables.

As baselines, we compare against leading global weather forecasting GNNs; methods such as GraphCast \cite{lam2023graphcast}, and hierarchical architectures, namely GraphFM and its ensemble variant GraphEFM \cite{oskarsson2024probabilistic}. All models incorporate solar radiation as a forcing variable and natively occupy the $5.625^\circ$ grid. The Flow models, however, apply max-pooling to acquire solar radiation across multiple scales. Implementation details are in the Appendix. Through pretrained weights, as described in Figure \ref{fig:arch_fig}, HiFlowCast is trained for one epoch on a single A100 GPU with 40 GB of RAM. In contrast, complete model training requires roughly $100$-$200$ epochs, which is approximately $100\times$ more expensive to train in terms of GPU clock time.

\begin{figure*}[t]
    \centering

    \begingroup
    \setlength{\abovecaptionskip}{3pt}
    \setlength{\belowcaptionskip}{3pt}
    \setlength{\dbltextfloatsep}{8pt}
    \setlength{\dblfloatsep}{8pt}

    \begin{subfigure}[t]{0.85\linewidth}
        \centering
        \includegraphics[width=\linewidth]{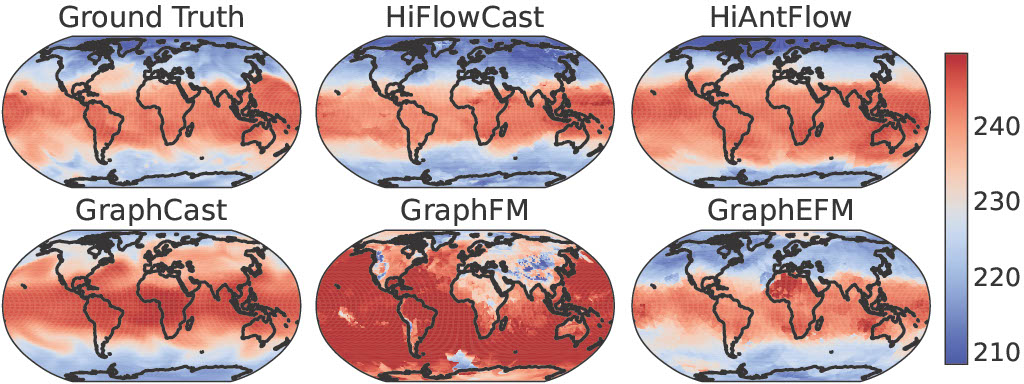}
        \vspace{2pt}
        \caption{Atmospheric temperature at a pressure level of $500$hPA after a 13-day lead time. HiFlowCast and HiAntFlow maintain global circulation, unlike baselines that collapse in equatorial heat transport.}
        \label{fig:forecast_fields}
    \end{subfigure}

    \begin{subfigure}[t]{0.85\linewidth}
        \centering
        \includegraphics[width=\linewidth]{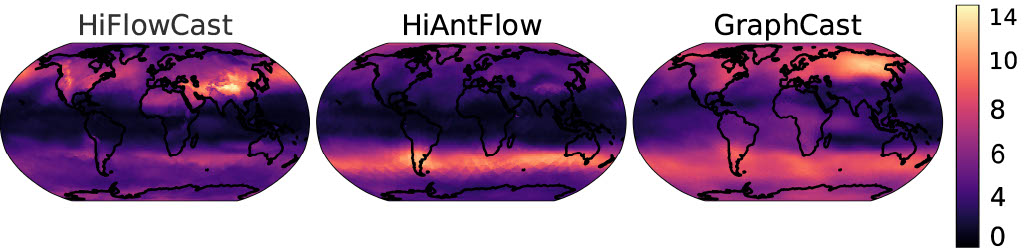}
        \vspace{2pt}
        \caption{Average MAE of temperature across starting times at a 10-day lead.}
        \label{fig:mae_map}
    \end{subfigure}

    \caption{Comparison of temperature forecasts at a 10-day lead time. HiFlowCast and HiAntFlow display high-fidelity projections over the rollout window.}
    \label{fig:lead_and_fields}
    \endgroup
\end{figure*}

\subsection{Long Lead Time Analysis}

Robust planning and analysis require foresight that extends beyond the immediate future. While approaches such as nowcasting \cite{ravuri2021dgmr, zhang2023nowcastnet} are valuable for capturing fine-grained details of near-term events, they fall short of identifying the critical signals that reveal broader system dynamics. For these applications, projections must remain faithful to ground truth observations and underlying physics principles. Consequently, we systematically evaluate model performance over a lead time of up to $13$-days in the future. This long lead time remains challenging for many existing methods \cite{chen2023long}.  

\begin{figure}[t]
    \centering
    \begin{minipage}{\linewidth}
        \centering
        \includegraphics[width=0.95\linewidth]{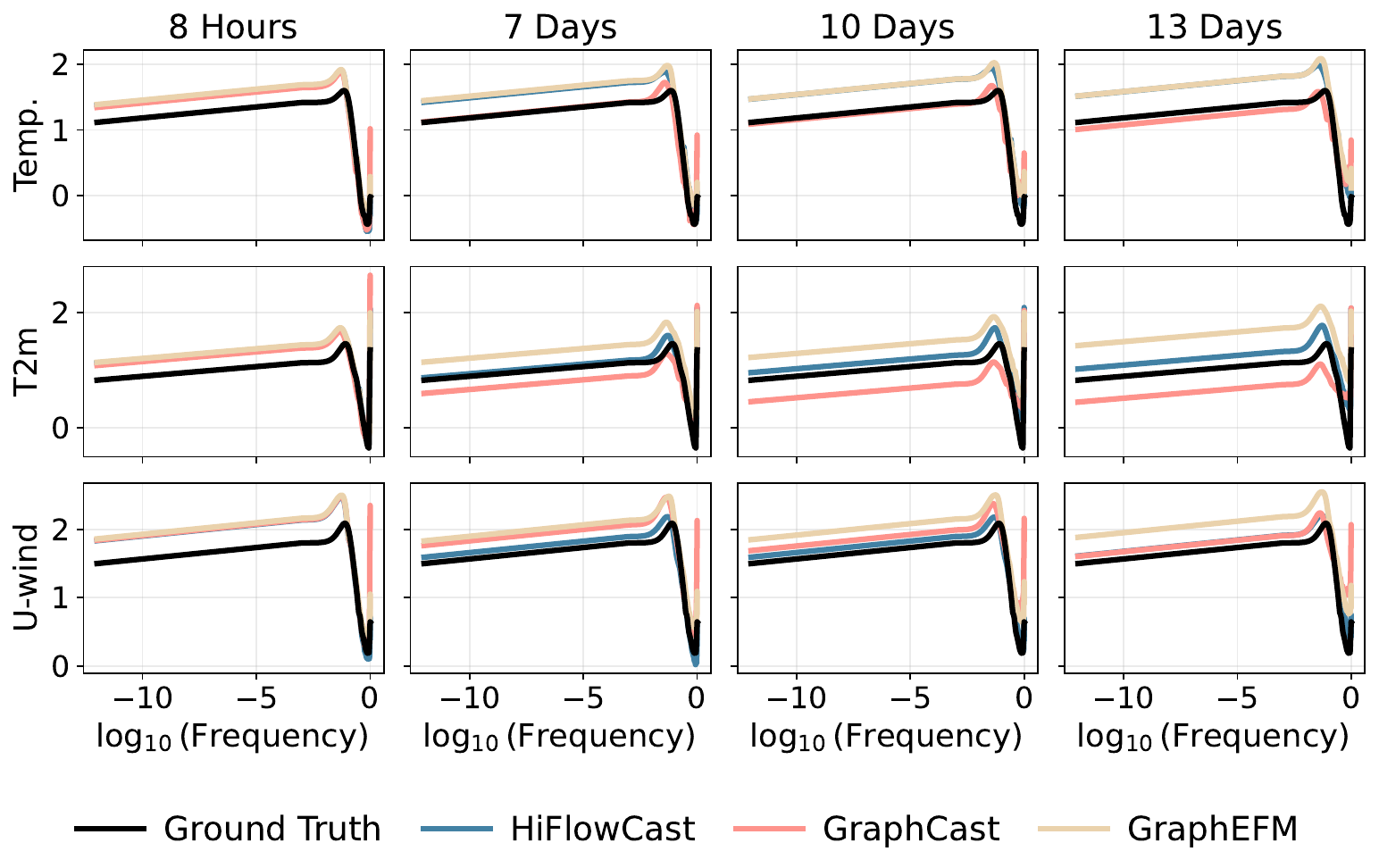}
        \caption{Power spectral density of projections over 8-hour and 13-day rollout windows. Proximity to the ground truth indicates how well models reproduce its detail.}
        \label{fig:smoothness_freq}
    \end{minipage}
\end{figure}

{\bf Sample performance on temperature.} Atmospheric temperature is a key variable for assessing climatological events such as heatwaves, and its importance has grown under the effects of global warming. Figure~\ref{fig:forecast_fields} presents projections at a 13-day lead time. GraphFM illustrates the compounding error typical of sequential forecasting models, leading to rapid collapse in the fidelity of forecasts. GraphCast, in contrast, only partially violates physical laws, most notably equatorial heat transport in the Northern Hemisphere. The Flow models preserve these transport laws and deliver higher-fidelity projections. GraphEFM remains competitive with the Flow models, though it tends to overestimate around Africa and underestimate regions across the Pacific. In the appendices, we provide additional example projections. 

Figure~\ref{fig:mae_map} reports MAE, averaged across lead and starting times, to highlight spatial bias between our HGNN models and the multi-mesh GNN, GraphCast. Our findings are consistent with the earlier observations surrounding heat transport around the poles. Specifically, GraphCast exhibits elevated errors along both poles of the equator. In contrast, the Flow models localize error to a single pole: HiAntFlow shows higher MAE in the Southern Hemisphere, whereas HiFlowCast shows the highest error in the Northern Hemisphere.

{\bf Frequency-Domain Smoothness Analysis.} As forecasts extend, they progressively lose high-frequency components, which are critical for maintaining stability at long lead times \cite{lippe2023pde}. To quantify the extent to which models preserve information across spatial scales, we compute the power spectral density (PSD) of forecasts for surface and atmospheric temperature and wind variables. Figure \ref{fig:smoothness_freq} presents the results. 

At short horizons (8 hours), all models remain close to the ground truth. By 7 days, however, GraphCast and GraphEFM diverge from the ground truth in surface temperature, reflecting a loss of fidelity. This degradation intensifies at 10 days and becomes pronounced at 13 days, where GraphCast and GraphEFM deviate substantially. In contrast, HiFlowCast maintains proximity to the ground truth throughout, preserving high-frequency structure over long lead times. We also observe that the PSD for HiFlowCast and GraphCast over meridional wind speed exhibit a high fidelity with respect to their low frequency components, while for temperature, GraphEFM and HiFlowCast are comparable. Note, however, this does not describe the alignment between model projections and intricate spatial patterns in this field.

\begin{figure*}[t]
    \centering
    
    \begin{minipage}{\linewidth}
        \centering
        \captionof{table}{Performance comparison at 1, 10, and final lead time (13 days). 
        Lower values indicate higher-fidelity projections. The best results are in bold, while the second best are underlined.}
        \vspace{10pt}
        \label{tab:metrics_1_10_last}
        \setlength{\tabcolsep}{7pt}
        \renewcommand{\arraystretch}{1.05}
        {\fontsize{7.7}{9}\selectfont
        \begin{tabular}{lcccccc}
            \toprule
            \multicolumn{7}{c}{\textbf{1-day Lead Time}} \\
            \midrule
            \textbf{Metric} & GraphFM & GraphCast & GraphEFM & HiFlowCast & HiAntFlow & \textbf{Gain}\\
            \midrule
            RMSE & 0.4989 & 0.4345 & 0.4576 & \textbf{0.4295} & \underline{0.4545} & \textcolor{gainGreen}{\textbf{+6.0\%}} \\
            MAE  & 0.3383 & 0.2985 & 0.3102 & \textbf{0.2925} & \underline{0.3062} & \textcolor{gainGreen}{\textbf{+5.7\%}} \\
            CRPS & 0.4725 & 0.4579 & 0.4628 & \textbf{0.4548} & \underline{0.4619} & \textcolor{gainGreen}{\textbf{+1.7\%}} \\
            \midrule
            \multicolumn{7}{c}{\textbf{10-day Lead Time}} \\
            \midrule
            RMSE & 1.0155 & 0.9636 & \underline{0.8323} & \textbf{0.8013} & 0.8390 & \textcolor{gainGreen}{\textbf{+3.7\%}} \\
            MAE  & 0.7409 & 0.6999 & \underline{0.5849} & \textbf{0.5678} & 0.5862 & \textcolor{gainGreen}{\textbf{+2.9\%}} \\
            CRPS & 0.6983 & 0.6701 & 0.6035 & \textbf{0.5815} & \underline{0.6026} & \textcolor{gainGreen}{\textbf{+3.6\%}} \\
            \midrule
            \multicolumn{7}{c}{\textbf{13-day Lead Time}} \\
            \midrule
            RMSE & 1.0844 & 1.0243 & \underline{0.8811} & \textbf{0.8338} & 1.0103 & \textcolor{gainGreen}{\textbf{+5.4\%}} \\
            MAE  & 0.7983 & 0.7495 & 0.6205 & \textbf{0.5935} & \underline{0.6022} & \textcolor{gainGreen}{\textbf{+4.4\%}} \\
            CRPS & 0.7397 & 0.7072 & 0.6269 & \textbf{0.5953} & \underline{0.6139} & \textcolor{gainGreen}{\textbf{+5.1\%}} \\
            \bottomrule
        \end{tabular}
        }
    \end{minipage}
    
\end{figure*}

\textbf{Error metrics over the entire dataset.} We evaluate model performance across all $13$ pressure levels and $11$ variables, normalizing inputs with $z$-score normalization so that each variable lies on the same scale. Table \ref{tab:metrics_1_10_last} shows our results. At short lead times, HiFlowCast achieves the lowest errors across all metrics, reducing RMSE and MAE by 6.0\% and 5.7\% over HiAntFlow, the second-best model at a 1-day lead time. These gains persist with a forecast horizon of 10 days; HiFlowCast lowers RMSE by 3.7\% and CRPS by 3.6\%, while at 13 days it delivers the most substantial improvements, reducing RMSE, MAE, and CRPS by 5.4\%, 4.4\%, and 5.1\%, respectively. HiAntFlow consistently ranks second, highlighting the robustness of the multiscale prediction framework, though it frequently underperforms with respect to the deterministic variant. The Appendices show extended results across all lead times and a per-variable analysis over all three metrics. 
 
\begin{figure*}[t]
    \centering

    \captionsetup[subtable]{justification=centering,singlelinecheck=false}
    \captionsetup[subfigure]{justification=centering,singlelinecheck=false}

    \begin{minipage}{\linewidth}
        \centering
        \captionof{table}{MAE at 13-day lead time for 1st and 99th quantile extremes. 
        Lower is better. Gains are relative to the best non-Flow baseline.}
        \vspace{10pt}
        \label{tab:extremes_subfig}
        \setlength{\tabcolsep}{6.7pt}
        \renewcommand{\arraystretch}{1.05}
        {\fontsize{7.7}{9}\selectfont
        \begin{tabular}{lcccccc}
            \toprule
            \multicolumn{7}{c}{\textbf{1st Quantile Extremes}} \\
            \midrule
            \textbf{Variable} & GraphFM & GraphCast & GraphEFM & HiFlowCast & HiAntFlow & \textbf{Gain} \\
            \midrule
            Temp.        & 18.81 & 8.74 & \underline{4.91} & \textbf{4.69} & 6.10 & \textcolor{gainGreen}{\textbf{+4.5\%}} \\
            U-wind       & 44.22 & 10.51 & \underline{10.50} & 11.28 & \textbf{9.97} & \textcolor{gainGreen}{\textbf{+5.0\%}} \\
            V-wind       & 19.53 & 9.47 & \underline{8.78} & 9.73 & \textbf{8.15} & \textcolor{gainGreen}{\textbf{+7.2\%}} \\
            Geopotential & 3668.24 & 2393.33 & \textbf{1173.81} & 1372.83 & 1420.96 & \textcolor{lossRed}{\textbf{--16.9\%}} \\
            \midrule
            \multicolumn{7}{c}{\textbf{99th Quantile Extremes}} \\
            \midrule
            Temp.        & 18.44 & 9.82 & \underline{6.07} & \textbf{5.78} & 7.12 & \textcolor{gainGreen}{\textbf{+4.8\%}} \\
            U-wind       & 45.31 & 10.54 & \underline{9.52} & 10.89 & \textbf{9.50} & \textcolor{gainGreen}{\textbf{+0.3\%}} \\
            V-wind       & 19.00 & 8.69 & \underline{8.33} & 9.19 & \textbf{7.61} & \textcolor{gainGreen}{\textbf{+8.6\%}} \\
            Geopotential & 3263.43 & 2773.87 & \underline{1593.56} & \textbf{1459.50} & 1698.89 & \textcolor{gainGreen}{\textbf{+8.4\%}} \\
            \bottomrule
        \end{tabular}
        }
    \end{minipage}

    \vspace{1.2em} 

    \begin{subfigure}[t]{\linewidth}
        \centering
        \includegraphics[width=\linewidth]{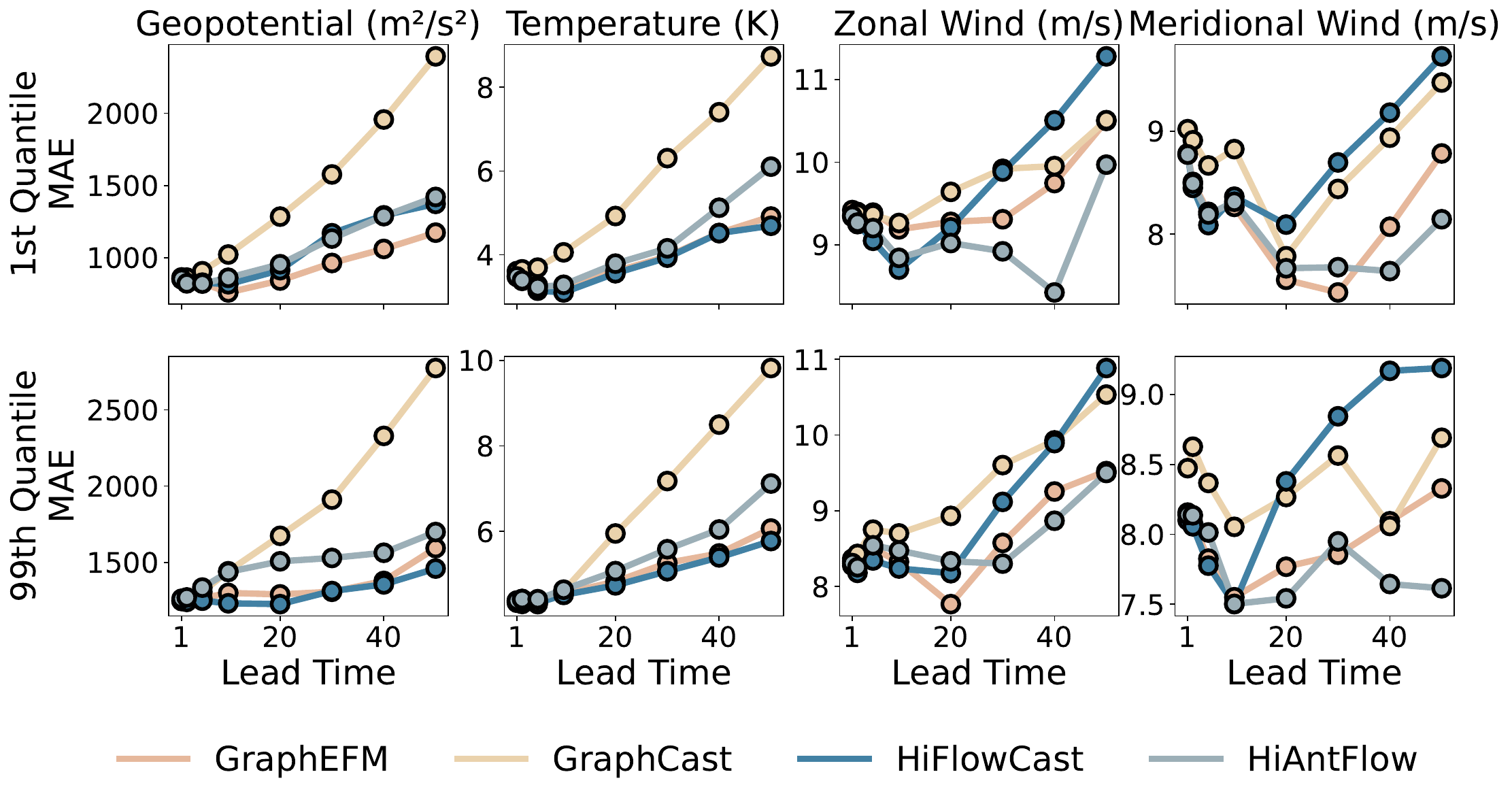}
        \caption{Quantile extreme predictions across lead times.}
        \label{fig:gabrielle_extremes}
    \end{subfigure}

    \caption{Cyclone Gabrielle: (a) Mean absolute error at 13-day lead time for quantile extremes across variables, including geopotential; (b) Predicted quantile extremes across lead times.}
    \label{fig:gabrielle_overview}
\end{figure*}

\subsection{Analysis over Extreme Events}

Extreme events are the most impactful weather scenarios to forecast. Their effects are both immediate and severe. Cyclones, for example, can devastate infrastructure across vast regions, while heatwaves, occurring with increasing frequency in a warming climate, pose escalating risks to human health and ecosystems. Our analysis, therefore, centers on these high-impact events, evaluating the capacity of the Flow models to anticipate their occurrence and intensity.

{\bf Long-lead projections over extremes.} To assess performance under rare yet critical conditions, we evaluate samples near the ground-truth distribution's $1$st and $99$th quantiles. We then unroll forecasts up to a $13$-day lead time and report averages over atmospheric temperature, zonal and meridional wind speeds, and geopotential.  

Figure~\ref{fig:gabrielle_extremes} summarizes the mean absolute error (MAE) across lead times, averaged over initialization dates, while Table~\ref{tab:extremes_subfig} reports values at the $13$-day horizon. HiFlowCast improves performance across both tails of temperature extremes, reducing error by $4.5\%$ and $4.8\%$ at the $1$st and $99$th quantiles, respectively. It also achieves an $8.4\%$ gain on $99$th-quantile geopotential extremes. These fields, which exhibit limited variability in sample complexity, are less affected by the single-outcome bias of deterministic forecasts. By contrast, zonal and meridional wind extremes present higher variability, and both HiFlowCast and GraphCast show a sharp error increase beyond $10$–$20$ days. Ensemble models mitigate this instability, reflecting the benefit of capturing diverse plausible outcomes. In particular, HiAntFlow yields the lowest errors for zonal wind, improving by $7.2\%$ and $8.6\%$ at the $1$st and $99$th quantiles. GraphEFM remains competitive on $99$th-quantile meridional wind, achieving the strongest performance on $1$st-quantile geopotential extremes by approximately $17\%$.

\section{Discussion}

We introduce the Flow models, HiFlowCast and its ensemble variant HiAntFlow, which integrate solar radiation fields into a multiscale prediction framework. By capturing the influence of radiation across spatio-temporal resolutions, these models address the loss of global-scale physics, a key limitation of current HGNNs, by embedding forcing terms across multiple scales. Rigorous evaluation shows that the Flow models markedly improve stability in projection fidelity for variables directly governed by radiation forcing, including temperature and geopotential. This gain arises because solar radiation modulates surface heating, which drives atmospheric and surface temperatures, air density, and zonal and meridional wind velocity fields. Extending the framework to incorporate additional PDE-based forcings, such as ocean currents or cloud-top cooling physics, would provide stricter constraints on model trajectories, potentially involving more intricate methodologies. Notably, at the 99th quantile, the Flow models achieve substantial reductions in MAE for temperature, zonal wind speed, and geopotential; variables critical for anticipating high-impact stress events such as heatwaves and cyclones.

Our approach also reduces computational cost. By exploiting transfer learning and the inherent generalizability of network components, a property akin to the universality of transformer models \cite{yun2019transformers}, we limit training to a single epoch. This strategy substantially lowers total GPU clock time during training, offering an alternative to prolonged regimes and broadening access to graph-based machine learning for global climate research. Specifically, at least when interaction networks connect graphs in the hierarchy, the space of possible operations comprises a small set of mappings. Existing components can thereby assist in building many unexplored HGNN architectures. Additionally, in the appendices, we provide a table of carbon emissions incurred during the evaluation stage, along with an asymptotic analysis. We show that the cost of additional operations introduced by the Flow models is bounded by the cost of evaluating base-level interaction networks.

HiFlowCast requires each output module to generate physically consistent projections to assemble a coherent forecast across scales. While this assumption holds at short lead times, the experiments in Figure~\ref{fig:gabrielle_extremes} demonstrate that it breaks down at longer horizons. When faced with out-of-distribution or limited samples, HiFlowCast struggles to maintain physical fidelity across its multiple predictions. HiAntFlow mitigates this limitation by averaging over multiple outcomes, which stabilizes projections but incurs a considerable computational cost from sampling.    









\clearpage
\bibliography{sn-bibliography}
\clearpage

\begin{appendices}

\section{Methodological Foundations}

\subsection{Problem Setting}

We formalize the global weather forecasting task and introduce the graph-based techniques used in this work. 

\subsection{Global Weather Forecasting}

Global weather forecasting is typically a \textit{sequential forecasting} problem: given the state $X^t$ at time $t$, the goal is to predict the next state $X^{t+1}$. In many cases, $X^t$ is only available as an initial ground-truth observation at $t=0$. Specific scenarios, however, provide an additional \textit{forcing state} $F^t$ at each time step, allowing the forecasting model $\mathbf{G}$ to constrain projections according to latent physical processes. This mapping is defined as:
\begin{equation}
    X^{t+1} = \mathbf{G}(X^t, F^t).
\end{equation}
When $X^t$ includes multiple climate variables, such as temperature, humidity, and precipitation at fixed atmospheric levels, the task becomes \textit{multivariate}. 

\subsubsection{Multi-mesh and Hierarchical Graphs}

\citet{lam2023graphcast} introduced the multi-mesh graph $\mathcal{G}_{M} = (\mathcal{V}, \mathcal{E}_1 \cup \dots \cup \mathcal{E}_N)$, where $\mathcal{E}_k$ denotes edges at resolution $k$ over the same node set $\mathcal{V}$. While multi-mesh graphs capture multiscale interactions, they lack an explicit mechanism for modelling dependencies between adjacent scales. As a result, oversquashing of node features can obscure cross-scale process dynamics. To overcome this limitation, we follow \citet{oskarsson2024probabilistic}, constructing a hierarchical graph family that explicitly couples information between scales. We define bipartite traversal graphs as:
\begin{equation}
    \mathcal{G}_{i \mapsto j} = (\mathcal{V}_i \cup \mathcal{V}_j, \mathcal{E}_{i \mapsto j}), \quad i,j \in [1,K],
\end{equation}
where the node sets $\mathcal{V}_i$ and $\mathcal{V}_j$ belong to mesh-graphs at adjacent resolutions, $\mathcal{G}_i = (\mathcal{V}_i, \mathcal{E}_i)$ and $\mathcal{G}_j = (\mathcal{V}_j, \mathcal{E}_j)$. When $i > j$, then $\mathcal{G}_{i\mapsto j}$ defines downward hierarchical traversal, while $i < j$ corresponds to upward traversal. This information flows in directions to learn local phenomena from global trends, and vice versa. However, when $i =j$, $\mathcal{G}_{i \mapsto i}$ is a bipartite graph that describes internal dynamics occurring on scale $i$.

The cross-scale edges $\mathcal{E}_{i \mapsto j}$ explicitly couple these levels, enabling bidirectional transfer of information, specifically via setting $\mathcal{E}_{i \mapsto j} \subseteq \mathcal{V}_i \times \mathcal{V}_j$. This hierarchical design forms a pyramid of interconnected processes. See Figure \ref{fig:arch_fig} for an illustration.

\subsubsection{Interaction Networks}  

Interaction networks~\cite{battaglia2016interaction, lam2023graphcast} learn dependencies between processes by modelling their pairwise interactions. HiFlowCast leverages an interaction network $\mathbf{M}_{i \mapsto j}$ to capture the relationships encoded in $\mathcal{G}_{i \mapsto j}$, for example, the macroscale physics driving cyclone formation and the local-scale phenomena that form as a consequence. When $i = j$, $\mathbf{M}_{i \mapsto j}$ models intra-scale dynamics within $\mathcal{G}_{i \mapsto i}$, and when $i \neq j$, it captures the cross-scale dynamics between processes that $\mathcal{G}_{i \mapsto j}$ connects. Interaction networks capture these internal and external processes by learning edge weights $W_{i \mapsto j}$:
\begin{equation}
    X^t_j \mapsto \mathbf{M}_{i \mapsto j}(X^t_i, X^t_j, W_{i \mapsto j}).
    \label{eq:int_map}
\end{equation}
Equation~\ref{eq:int_map} then updates hidden features at level $j$ in the hierarchy. We initialize $X^t_j$ as the zero tensor before the upward hierarchical pass.

\subsection{The Flow Models}

HiFlowCast is an HGNN designed to forecast climate dynamics across multiple spatio-temporal scales. The Flow framework explicitly couples global and local processes, while embedding physical constraints onto projections that span various resolutions. These mechanisms aim to enable HiFlowCast to preserve coherence with global and regional-scale physical processes. Furthermore, HiFlowCast admits a natural extension, HiAntFlow, an ensemble variant that captures the intrinsic chaotic sensitivity of the climate system to initial conditions.

\subsubsection{Retaining Multiscale Trends}

A central challenge in HGNN models is preserving global-scale information during downward traversal. While passing information downwards to the highest resolution ideally embeds latent physics into local representations, finite buffer sizes and nonlinear transformations often cause information loss \cite{topping2022understanding}. Such forgetting limits the effective forecasting horizon by compounding error over time. Adding residual connections between dynamics occurring over $\mathcal{G}_{i \mapsto j}$ and $\mathcal{G}_{j \mapsto i}$, for $i \neq j$, mitigates these effects to a degree. However, they fail to capture the joint dependency between upward and downward information flows, limiting the model's ability to exploit local phenomena contextualizing learned global trends.

HiFlowCast introduces a Latent-Memory-Retention module in the downward pass to address these issues. The module updates local features by jointly processing upward and downward flows, contextualizing global dynamics regarding local phenomena. Specifically, it enables concurrent modeling by projecting the memory buffer onto the next lower hierarchical level. We define a global-to-local network $\mathbf{B}_{k+1 \mapsto k}$ that constructs a local-scale prior $U^t_k$. A scale-preserving map $\mathbf{D}_{k \mapsto k}$ then predicts the memory buffer $H^t_k$ from $U^t_k$ and $X^t_k$:
\begin{equation}
    U^t_k = \mathbf{B}_{k+1 \mapsto k}(H_{k+1}^t, \mathbf{0}, W_{k \mapsto k+1}^t),
    \label{eq:utk}
\end{equation}
\begin{equation}
    H_k^t = \mathbf{D}_{k \mapsto k}(U_k^t, X_k^t, \widetilde{W}^t_{k \mapsto k}),
    \label{eq:htk}
\end{equation}
where $W^t_{k\mapsto k+1}$ and $\widetilde{W}^t_{k \mapsto k}$ are learnable edge weights and $\mathbf{0}$ is the zero tensor. We implement $\mathbf{D}_{k \mapsto k}$ as a residual network with $S$ layers. Specifically, at each layer the network updates $\widetilde{W}^t_{k \mapsto k}$, allowing for learning of intricate internal processes.  

\subsubsection{Embedding Physics into Predictions}

Global-scale dynamics drive local phenomena, with large-scale physical processes shaping high-resolution behaviour. Yet, global trends are difficult to preserve when repeatedly compressed through downward transformations \cite{dwivedi2022gnn,topping2022understanding,alon2021on}.

Topology-free multiscale models \cite{challu2023nhits,oreshkin2019n} negate these issues by directly accessing multi-level features to form multiscale predictions. Inspired by these innovations, HiFlowCast introduces a Latent-to-Physics branch, a dedicated pathway that injects physics constraints derived from PDE solution fields directly into latent memory buffer states $H^t_k$. Specifically, as shown in Figure~\ref{fig:arch_fig}, the forcing input $F^t$ is max-pooled to a coarse representation $F^t_i$, which fuses with memory buffer features through a mesh-to-grid interaction network $\mathbf{P}_{k \mapsto k}$. Concretely, the mapping can be written as: 
\begin{equation}
    F^t_i = \mathrm{MaxPool}^{(i)}(F^t),
\end{equation}
\begin{equation}
    \hat{Y}^t_i = \sum_{m \leq i} \mathbf{P}_{m\mapsto m}(H^t_m, F^t_m).
    \label{eq:hiflowcast_out}
\end{equation}
The design of HiFlowCast ensures that multiscale predictions remain consistent with physics described by the solution fields. Furthermore, these global-scale predictions constrain regional forecasts to align with predictions originating from higher levels. We train with a mean squared error loss over the summation of outputs from each layer.

\subsubsection{Deriving an Ensemble Variant}

To account for the inherent chaos of the climate system \cite{lam2023graphcast}, we extend HiFlowCast into an ensemble formulation, HiAntFlow. By explicitly modelling uncertainty, the ensemble mitigates spurious trajectories produced by its deterministic backbone, suppressing unlikely realizations while amplifying robust projections representing likely outcomes. The resulting distributions concentrate around the most plausible outcomes, which are generally preferable in forecasting applications with a vast space for possible scenarios. 

Our formulation for HiAntFlow follows closely that proposed in \citet{oskarsson2024probabilistic}, which samples latent variables from a normal distribution defined over the lowest level feature-space. In particular, the sampling may be written as: 
\begin{equation}
    Z \sim \mathcal{N}(\mu(X^t), \sigma I),
\end{equation}
where $\mu$ is a latent mean extraction branch described in Figure \ref{fig:arch_fig} of the main paper, and $\sigma$ the chosen diversity hyperparameter. We take the expectation of the marginal distribution over the latent variables $Z$ to calculate the final output of HiAntFlow:
\begin{equation}
\hat{Y}_i^t = \sum_{m \leq i}\mathbb{E}_{Z}\Big[\mathbf{P}_{m \mapsto m}(H^t_m, F^t_m, Z)\Big],
\end{equation}
where the mesh-to-grid interaction network $\mathbf{P}_{m\mapsto m}$ is extended to incorporate the influence of the latent variables, in practice, we approximate the expectation by computing the empirical mean of the marginal distribution with respect to $Z$.

\subsection{Asymptotic Analysis}

Global climate datasets span long temporal scales and are often at a high spatial resolution, making scalability a central concern for training and evaluation. We analyze the asymptotic computational complexity of the Flow models over the latter case. In particular, we focus on the additional cost introduced by operations during the downward traversal incurred via the Latent-Memory-Retention and Latent-to-Physics branches (Figure~\ref{fig:arch_fig}).  

Downward traversal consists of the GNN mappings in Equations~\ref{eq:utk}, \ref{eq:htk}, and \ref{eq:hiflowcast_out}. Let $\mathcal{C}: \Omega(X^t, F^t)\to\mathbb{R}$ denote the cost functional over the space of interaction networks $\Omega(X^t, F^t)$ that take projection $X^t$ and forcing field $F^t$ as input at time $t$. For a hierarchy of depth $K$, where Equation~\ref{eq:htk} implements an $S$-layer residual network, the total cost of downward traversal is:
\begin{align}
    \sum_{k=1}^K \Big( C(\mathbf{B}_{k+1\to k}) + S\cdot C(\mathbf{D}_{k\to k}) + \mathcal{C}(\mathbf{P}_{k\to k}) \Big).
\end{align}
Since the lowest level typically dominates (due to the largest graphs), this scales as
\begin{align}
    \mathcal{O}\!\Big(K \cdot \big(C(\mathbf{B}_{2\to 1}) + S\cdot C(\mathbf{D}_{1\to 1}) + C(\mathbf{P}_{1\to 1})\big)\Big).
\end{align}

By contrast, a lightweight hierarchical variant without memory buffers, multiple mesh-to-grid mappings, or mesh-to-mesh mappings is bounded above by a cost of
\begin{align}
    \mathcal{O}\!\Big(C(\mathbf{P}_{1\to 1}) + K\cdot C(\mathbf{B}_{2\to 1})\Big),
\end{align}
which represents a bound on the minimal cost of an HGNN, since at least one projection and one downward mapping is unavoidable.

The ratio of Flow to lightweight complexity is therefore
\begin{align}
    \mathcal{O}\!\Bigg(
    \frac{K\cdot\big(C(\mathbf{B}_{2\to 1}) + S\cdot C(\mathbf{D}_{1\to 1}) + C(\mathbf{P}_{1\to 1})\big)}
    {C(\mathbf{P}_{1\to 1}) + K\cdot C(\mathbf{B}_{2\to 1})}
    \Bigg).
\end{align}
As $K$ grows large, this ratio converges to
\begin{align}
    \mathcal{O}\!\Bigg(1 + \frac{S\cdot C(\mathbf{D}_{1\to 1}) + C(\mathbf{P}_{1\to 1})}{C(\mathbf{B}_{2\to 1})}\Bigg).
\end{align}

If we further assume $C(f)= \delta\cdot|f|$, where $|f|$ denotes the parameter count of network $f$, and $\delta \in \mathbb{R}^+$, then the Flow models are approximately
\begin{align}
    \Bigg(1+\frac{S\cdot \eta_D + 1}{\eta_B}\Bigg)\times
\end{align}
more expensive than the lightweight variant if
$\eta_B|\mathbf{B}_{2\to 1}| = |\mathbf{P}_{1\to 1}| = \eta_D|\mathbf{D}_{1\to 1}|$ for constants $\eta_B,\eta_D\in\mathbb{R}^+$.  

Thus, the overhead of the Flow models remains asymptotically constant regarding hierarchy depth and scales, with only the relative parameter sizes between the memory buffer and downward mesh-to-mesh.

\subsubsection{Related Work}

{\bf Learning Multiscale Trends.} Hierarchical architectures have emerged as powerful tools for forecasting across diverse domains \cite{wu2023hiformer,qin2023scaleformer,nie2023patchtst,cao2024hierarchical,cini2023graph}. These models uncover local and global dynamics by capturing relationships at multiple resolutions. Graph-based approaches extend this multiscale learning principle by overlaying a graph structure onto a time series \cite{cini2023graph, cao2024hierarchical} and pooling to obtain coarse representations of global interactions. The construction of coarse graphs yields macro-level summaries of spatio-temporal dependencies.
In contrast, transformer-based methods, which lack explicit topological constraints, implicitly aim to infer such multiscale trends. Autoformer \cite{wu2021autoformer}, for example, decomposes input signals into seasonal and local components, exposing the model to critical underlying drivers. Recent comparable models, such as TimeMixer and its extension TimeMixer++ \cite{wang2024timemixer, wang2024timemixer++}, further refine seasonal decomposition by explicitly mixing multiscale seasonal and local trends. Meanwhile, other approaches \cite{challu2023nhits,oreshkin2019n} design explicit multi-resolution signal decompositions to enhance forecasting. However, topology-free methods often fail to reflect the intricate dependencies that govern the inherent intricacies of global climate systems. Accurate long-lead climate projections demand explicit modeling of interactions between large-scale drivers, which purely statistical trend decompositions cannot guarantee.  

{\bf Global Weather \& Climate Forecasting.} Numerical climate models \cite{bauer2015quiet, schneider2017earth, palmer2000predictability} remain the foundation of long-range forecasting, evolving states forward in time by solving coupled Partial Differential Equations (PDEs). Their high-fidelity stems from embedding physical constraints directly into each rollout step, enabling stable, long lead time projections \cite{reichstein2019deep, dueben2018challenges} that integrate observational data. 

In contrast, data-driven approaches seek to emulate these governing processes by learning from historical data. For instance, NeuralGCM \cite{kochkov2024neural} combines atmospheric circulation constraints into a data-driven model, preserving these physical aspects during the unrolling of projections even up to a long lead time. Other recent advances employ transformer-based architectures. FourCastNet \cite{pathak2022fourcastnet} leverages Fourier neural operators to capture global-scale atmospheric dynamics, achieving competitive long-range forecasts. Pangu-Weather \cite{bi2023pangu} utilizes a 3D Earth-specific transformer to represent multiscale atmospheric interactions and deliver state-of-the-art performance. Other approaches instead adopt graph-based formulations. GraphCast \cite{lam2023graphcast} employs GNN on a spherical mesh to represent Earth’s surface and achieve high accuracy at extended lead times. In contrast, Spherical Fourier Neural Operators \cite{bonev2023spherical} incorporate periodic constraints to learn stable dynamics directly on the sphere, further enhancing consistency with underlying physics occurring at the poles. 
 While some recent advances aim to capture multiscale trends, the lack of an explicit computational structure means they do not integrate physics across multiple spatio-temporal scales.

HiFlowCast builds on these ideas by embedding PDE solution fields directly into its multiscale GNN forecasting framework. Integrating these physics-oriented constraints aims to enhance the stability of long-lead projections, consequently extending the high-correlation range of forecasts.  

{\bf Direct Modeling with Graph Neural Networks.} GNNs \cite{kipf2016semi, wu2020comprehensive} have transformed diverse fields, from molecular discovery to forecasting \cite{bessadok2022graph, keisler2022forecasting, liao2021review, jiang2023graph, rahmani2023graph}. Through representing entities as nodes and their relationships as edges, GNNs naturally capture structured interactions \cite{bronstein2021geometric, bronstein2023topological}. This relational bias is particularly advantageous in weather forecasting, as atmospheric processes are inherently multivariate and coupled across diverse scales. For instance, GraphCast \cite{lam2023graphcast} demonstrates enhanced modeling capacity by projecting climate variables onto a multi-mesh graph, enabling multiscale interactions to occur on a single graph. Yet, highly internally coupled GNNs are prone to oversquashing \cite{barcelo2022graph}, a phenomenon where a substantial amount of information is compressed into fixed-size node embeddings \cite{dwivedi2022gnn,topping2022understanding,alon2021on}, limiting the ability of these GNNs to store and propagate information. More recent approaches \cite{oskarsson2024probabilistic} alleviate oversquashing by coupling graphs positioned in a hierarchy with interaction networks \cite{battaglia2016interaction}. These networks model trends occurring at different scales as mutually influential, yet separate, processes, elevating oversquashing. Still, hierarchical architectures often struggle to retain global-scale trends across depths \cite{ying2018hierarchical,zhang2021hierarchical, ma2019graph}. HiFlowCast addresses these issues by enforcing multiscale memory retention and embedding global solar radiation into its solution fields, strengthening the role of multiscale physics within unrolled projections.

\subsection{Metrics}
We evaluate model performance using Root Mean Squared Error (RMSE), Mean Absolute Error (MAE), and the Continuous Ranked Probability Score (CRPS). RMSE highlights large deviations by penalizing squared differences, making it sensitive to substantial forecast errors. MAE captures the average error magnitude, providing a robust view of overall accuracy. CRPS measures the quality of probabilistic forecasts by comparing the predicted distribution against the observed outcome, thereby assessing both sharpness and calibration. These metrics give complementary perspectives: RMSE emphasizes regions of high error, MAE reflects typical accuracy, and CRPS evaluates probabilistic reliability. Formally, given forecasts $\{\hat{y}_t\}_{t=1}^T$ and corresponding observations $\{y_t\}_{t=1}^T$, they are defined as:
\begin{align}
    \text{RMSE} &= \sqrt{\frac{1}{T}\sum_{t=1}^T (\hat{y}_t - y_t)^2}, \\
    \text{MAE}  &= \frac{1}{T}\sum_{t=1}^T |\hat{y}_t - y_t|, \\
    \text{CRPS}(F,y) &= \int_{-\infty}^{\infty} \left( F(x) - \mathbf{1}\{y \leq x\} \right)^2 dx,
\end{align}
where $F(x)$ is the predictive cumulative distribution function and $\mathbf{1}\{\cdot\}$ the indicator function that maps to $0$ if the argument is false, and $1$ otherwise.

In addition, we use the Power Spectral Density (PSD) to evaluate how forecasts degrade over lead time by measuring the loss of high-frequency content in their projections. PSD measures the quantity of a frequency within a particular sample and serves as a proxy for smoothness. Spectra aligned with the ground truth avoid spurious high-frequency noise while retaining detail. Formally, for a time series $y_t$, the PSD is:
\begin{align}
    P(\omega) = \frac{1}{T}\Bigg|\sum_{t=1}^T y_t e^{-i \omega t}\Bigg|^2,
\end{align}
where $\omega$ denotes frequency. PSD thus complements RMSE, MAE, and CRPS by capturing qualitative aspects beyond error magnitude, which is often driven by the precise placement of prediction patterns. 

\subsection{Implementation Details}

\definecolor{softgreen}{RGB}{34,139,34}   
\definecolor{softred}{RGB}{178,34,34}     

\newcommand{\tick}{\textcolor{softgreen}{\scalebox{1.1}{\ding{51}}}} 
\newcommand{\cross}{\textcolor{softred}{\scalebox{1.1}{\ding{55}}}} 

\begin{table}[t]
\centering
\caption{Initialization procedures for the Flow models. \textit{Pretrained} use existing weights, \textit{Transfer Learning} retain trainable gradients, \textit{Generalizability} have frozen gradients, and \textit{Initialized} start with near-zero weights. The lower block, \textit{Proposed Modules}, highlights new components introduced in this work.}

\begin{tabular}{lcccc}
\toprule 
\textbf{GNN Role} & \textbf{Pretrained} & \textbf{Transfer Learning} & \textbf{Generalizability} & \textbf{Initialized} \\
\midrule 
Upward Pass       & \tick & \cross & \tick & \cross \\
Downward Pass     & \tick & \cross & \tick & \cross \\
Ensemble Prior    & \tick & \cross & \tick & \cross \\
\midrule 
\multicolumn{5}{c}{\hspace{2em}Proposed Modules} \\
\midrule 
Memory-Buffer     & \tick & \tick & \cross & \cross \\
Physics Embedders & \cross & \cross & \cross & \tick \\
\bottomrule
\end{tabular}
\label{tab:network_status}
\end{table}

We outline key implementation details for HiFlowCast and HiAntFlow, along with the experimental setups, code origins, and computational resources used.

{\bf Pretrained weights \& architecture.} Figure~\ref{fig:arch_fig} illustrates the overall Flow model architecture. We freeze the interaction networks responsible for grid-to-mesh mapping and upward and downward traversal between hierarchical levels. During upward traversal, we omit scale-preserving mesh-to-mesh networks, as preliminary experiments showed reduced accuracy, while still requiring additional evaluations. The training procedures for the rest of our modules are displayed in Table \ref{tab:network_status}. In particular, we load and fine-tune the mesh-to-mesh networks used in the memory-buffer mappings, allowing gradients to flow so that the buffer adapts to the forecasting task, which differs from the original problem setting during its training phase. We randomly initialize all weights in the Latent-to-Physics module. To minimize additional expense, we employ a residual network with $S=3$ layers for the Latent-Retention branch.  

{\bf Computational resources.} Training and evaluation were performed on a single NVIDIA A100 GPU (40\,GB memory). Our ERA5 dataset amounts to roughly 1.5\, TB, requiring segmentation and incremental training. Our implementation builds on a fork of the repository from \citet{oskarsson2024probabilistic}, available publicly at \url{https://github.com/mllam/neural-lam.git}. 

\begin{table*}[t]
    \centering
    \caption{ERA5 variables used in our datasets.}
    \label{tab:era5_vars}
    \setlength{\tabcolsep}{8pt}
    \renewcommand{\arraystretch}{1.1}
    {\small
    \begin{tabular}{lll}
        \toprule
        \textbf{Type} & \textbf{Variable name} & \textbf{Short name} \\
        \midrule
        Atmospheric & Geopotential & z \\
        Atmospheric & Specific humidity & q \\
        Atmospheric & Temperature & t \\
        Atmospheric & U component of wind & u \\
        Atmospheric & V component of wind & v \\
        Atmospheric & Vertical velocity & w \\
        \midrule
        Single & 2 metre temperature & 2t \\
        Single & 10 metre u wind component & 10u \\
        Single & 10 metre v wind component & 10v \\
        Single & Mean sea level pressure & msl \\
        Single & Total precipitation (6hr) & tp \\
        \bottomrule
    \end{tabular}}
\end{table*}

\begin{table*}[t]
    \centering
    \caption{ERA5 standard pressure levels with approximate altitudes.}
    \label{tab:era5_pressure_levels}
    \setlength{\tabcolsep}{20pt}
    \renewcommand{\arraystretch}{1.1}
    {\small
    \begin{tabular}{ll}
        \toprule
        \textbf{Pressure (hPa)} & \textbf{Alt. (km)} \\
        \midrule
        1000 & ~0 \\
        925  & ~0.7 \\
        850  & ~1.5 \\
        700  & ~3.0 \\
        600  & ~4.5 \\
        500  & ~5.5 \\
        400  & ~7.0 \\
        300  & ~9.0 \\
        250  & ~10.5 \\
        200  & ~12 \\
        150  & ~14 \\
        100  & ~16 \\
        50   & ~20 \\
        \bottomrule
    \end{tabular}}
\end{table*}

{\bf ERA5 Dataset Variables.} Tables \ref{tab:era5_vars} and \ref{tab:era5_pressure_levels} summarize the ERA5 reanalysis variables and their atmospheric topology used in the study. In particular, atmospheric fields include temperature, zonal and meridional wind speeds, specific humidity, and vertical velocity, all taken at atmospheric levels at pressures ranging from $50$ hPa to $1000$ hPa. Furthermore, single-level fields cover 2-metre temperature, 10-metre winds, mean sea level pressure, and $6$ hour precipitation. For clarity, we accompany each field with its standard ERA5 abbreviation. Table \ref{tab:era5_pressure_levels} further outlines the specific pressure levels over which the atmospheric variables range, from the surface ($1000$ hPa, $\sim0$ km) to the lower stratosphere ($50$ hPa, $\sim20$ km), reflecting the vertical discretizations of the global climate system. Our selected variables enable consistent representation of both surface conditions and free-atmosphere dynamics.

\section{Additional Experiments}

This section presents supplementary experiments that aim to support the investigations carried out in the main paper. 

\begin{table*}[t]
    \centering
    \caption{Performance comparison across multiple lead times. Lower values indicate higher-fidelity projections. The best results are in bold, while the second best are underlined.}
    \label{tab:metrics_all}
    \setlength{\tabcolsep}{6.5pt}
    \renewcommand{\arraystretch}{1.0}
    {\fontsize{7.7}{9}\selectfont
    \begin{tabular}{lcccccc}
        \toprule
        \multicolumn{7}{c}{\textbf{4-hour Lead Time}} \\
        \midrule
        \textbf{Metric} & GraphFM & GraphCast & GraphEFM & HiFlowCast & HiAntFlow & \textbf{Gain} \\
        \midrule
        RMSE & 0.3410 & 0.3038 & \textbf{0.2649} & \underline{0.2662} & 0.2600 & \textcolor{lossRed}{\textbf{-0.5\%}} \\
        MAE  & 0.2308 & 0.2074 & \textbf{0.1778} & \underline{0.1801} & 0.1733 & \textcolor{lossRed}{\textbf{-2.5\%}} \\
        CRPS & 0.4380 & 0.4327 & \underline{0.4260} & \textbf{0.4258} & 0.4254 & \textcolor{gainGreen}{\textbf{+0.05\%}} \\
        \midrule
        \multicolumn{7}{c}{\textbf{8-hour Lead Time}} \\
        \midrule
        RMSE & 0.4053 & 0.3623 & \underline{0.3328} & \textbf{0.3229} & 0.3313 & \textcolor{gainGreen}{\textbf{+3.0\%}} \\
        MAE  & 0.2744 & 0.2441 & \underline{0.2221} & \textbf{0.2175} & 0.2223 & \textcolor{gainGreen}{\textbf{+2.1\%}} \\
        CRPS & 0.4501 & 0.4438 & \underline{0.4360} & \textbf{0.4339} & 0.4361 & \textcolor{gainGreen}{\textbf{+0.5\%}} \\
        \midrule
        \multicolumn{7}{c}{\textbf{3-day Lead Time}} \\
        \midrule
        RMSE & 0.6267 & 0.5562 & \underline{0.5841} & \textbf{0.5526} & 0.5845 & \textcolor{gainGreen}{\textbf{+0.7\%}} \\
        MAE  & 0.4338 & 0.3863 & \underline{0.4004} & \textbf{0.3787} & 0.3990 & \textcolor{gainGreen}{\textbf{+2.0\%}} \\
        CRPS & 0.5147 & 0.4918 & \underline{0.5006} & \textbf{0.4883} & 0.5003 & \textcolor{gainGreen}{\textbf{+2.5\%}} \\
        \midrule
        \multicolumn{7}{c}{\textbf{5-day Lead Time}} \\
        \midrule
        RMSE & 0.8057 & 0.7426 & \underline{0.7102} & \textbf{0.6929} & 0.7127 & \textcolor{gainGreen}{\textbf{+2.5\%}} \\
        MAE  & 0.5718 & 0.5288 & \underline{0.4934} & \textbf{0.4828} & 0.4928 & \textcolor{gainGreen}{\textbf{+2.1\%}} \\
        CRPS & 0.5888 & 0.5624 & \underline{0.5484} & \textbf{0.5380} & 0.5481 & \textcolor{gainGreen}{\textbf{+1.9\%}} \\
        \midrule
        \multicolumn{7}{c}{\textbf{8-day Lead Time}} \\
        \midrule
        RMSE & 0.9292 & 0.8779 & \underline{0.7785} & \textbf{0.7667} & 0.7857 & \textcolor{gainGreen}{\textbf{+1.5\%}} \\
        MAE  & 0.6700 & 0.6320 & \underline{0.5443} & \textbf{0.5377} & 0.5469 & \textcolor{gainGreen}{\textbf{+1.2\%}} \\
        CRPS & 0.6500 & 0.6242 & \underline{0.5783} & \textbf{0.5670} & 0.5789 & \textcolor{gainGreen}{\textbf{+2.0\%}} \\
        \midrule
        \multicolumn{7}{c}{\textbf{Average Across All Leads (4h, 8h, 1d, 3d, 5d, 8d, 10d, 13d)}} \\
        \midrule
        RMSE & 0.7870 & 0.7416 & \underline{0.6290} & \textbf{0.6082} & 0.6342 & \textcolor{gainGreen}{\textbf{+3.3\%}} \\
        MAE  & 0.5523 & 0.5103 & \underline{0.4565} & \textbf{0.4421} & 0.4656 & \textcolor{gainGreen}{\textbf{+3.2\%}} \\
        CRPS & 0.5977 & 0.5724 & \underline{0.5285} & \textbf{0.5151} & 0.5339 & \textcolor{gainGreen}{\textbf{+2.6\%}} \\
        \bottomrule
    \end{tabular}
    }
\end{table*}

\subsection{Other Lead Time Tables}

We extend our lead-time analysis by evaluating performance across additional lead times, ranging from $4$-hours to $13$-days into the future. Table~\ref{tab:metrics_all} reports these results. The Table also reports the average across all leads. At the shortest horizon (4 hours), HiAntFlow attains the lowest MAE ($2.5\%$ gain) and the strongest RMSE overall, while HiFlowCast achieves the best CRPS. At $8$ hours, HiFlowCast leads across all three metrics, reducing RMSE by $3.0\%$ and MAE by $2.1\%$ relative to the following best baseline. The advantage persists at $3$ and $5$ days, with HiFlowCast consistently outperforming GraphEFM, particularly in CRPS, where improvements reach $2.5\%$. By $8$ days, the Flow models continue to lead, with HiFlowCast lowering RMSE by $1.5\%$ and CRPS by $2.0\%$. Averaged across all horizons, HiFlowCast achieves the strongest overall performance, reducing RMSE by $3.3\%$, MAE by $3.2\%$, and CRPS by $2.6\%$. 

These results demonstrate that embedding multiscale physics yields stable accuracy across both short and long-range forecasts. At a very short lead time, however, GraphEFM emerges as the highest performing model, with most of the benefits of the Flow model only being seen around $10$ and $13$ days into the future. Otherwise, the gains across the three metrics tend to be smaller.

\subsection{Comparison with IFS Forecasts}

\begin{table}[t]
    \centering
    \caption{Flow models vs. \textsc{IFS HRes} at 10-day lead time. The best model is given in bold, the second best is underlined.}
    \label{tab:metrics_hres_flow}
    \setlength{\tabcolsep}{8pt}
    \renewcommand{\arraystretch}{1.0}
    {\fontsize{7.7}{9}\selectfont
    \begin{tabular}{lcccc}
        \toprule
        \multicolumn{5}{c}{\textbf{10-day Lead Time}} \\
        \midrule
        \textbf{Metric} & \textsc{IFS HRes} & HiFlowCast & HiAntFlow & \textbf{Gain} \\
        \midrule
        RMSE & 0.875 & \textbf{0.801} & \underline{0.872} & \textcolor{gainGreen}{\textbf{+8.4\%}} \\
        MAE  & 0.696 & \underline{0.567} & \textbf{0.565} & \textcolor{gainGreen}{\textbf{+18.7\%}} \\
        CRPS & {\bf 0.552} & \underline{0.581} & 0.590 & \textcolor{lossRed}{\textbf{-5.2\%}} \\
        \bottomrule
    \end{tabular}
    }
\end{table}

Table \ref{tab:metrics_hres_flow} compares the 10-day forecast performance of the operational IFS HRES model with our Flow-based architectures. Following the evaluation protocol of \citet{lam2023graphcast}, IFS HRES is assessed against its own operational analysis, whereas HiFlowCast and HiAntFlow are evaluated against ERA5 to ensure fairness under open reanalysis data. To allow direct comparison across metrics, all scores are z-normalized. Each model uses the same 13 pressure levels listed in Table \ref{tab:era5_pressure_levels} and are restricted to the standardized WeatherBench2 \cite{rasp2024WeatherBench2} variable set, which includes wind components, temperature, humidity, and surface pressure fields (a subset of Table \ref{tab:era5_vars}). Overall, the Flow-based models achieve lower RMSE and MAE than IFS HRES, indicating higher deterministic accuracy at extended lead times, while IFS HRES retains a slight advantage in probabilistic sharpness, reflected in its lower CRPS.


\subsection{Flops and Carbon Footprint}

\begin{table}[t]
\centering
\caption{Training cost and parameter comparison across forecasting models. Metrics report total runtime, average energy use, estimated carbon emissions over the full training schedule, and model size in millions of trainable parameters.}
\label{tab:model_comparison_eval}
\setlength{\tabcolsep}{6.5pt}
\renewcommand{\arraystretch}{1.0}
{\fontsize{7.7}{9}\selectfont
    \begin{tabular}{lccccc}
        \toprule
        Metric & HiAntFlow & GraphEFM & HiFlowCast & GraphFM & GraphCast \\
        \midrule
        Real time (h) & $0.556$ & $0.558$ & $0.058$ & $0.109$ & $0.032$ \\
        Energy (kWh)  & $0.94$  & $0.94$  & $0.067$ & $0.067$ & $0.044$ \\
        Emissions (kg CO$_2$) & $0.24$  & $0.24$  & $0.020$ & $0.022$ & $0.016$ \\
        \bottomrule
    \end{tabular}
}
\end{table}

Table \ref{tab:model_comparison_eval} summarizes performance statistics collected over one year of evaluation. Runtime is reported in hours, energy consumption in kilowatt-hours (kWh), and carbon emissions in kilograms of CO$_2$ using CodeCarbon \cite{benoit_courty_2024_11171501}. Among all evaluated systems, HiFlowCast demonstrates the highest computational efficiency after GraphCast, achieving comparable accuracy with only a fraction of the runtime and energy required by the other hierarchical baselines. In contrast, HiAntFlow and GraphEFM exhibit substantially higher computational costs, underscoring the heavier demands of ensemble-based and deterministic HGNN architectures.

\subsection{Disaggregated Analysis over Variables}

\begin{figure*}
    \centering
    \includegraphics[width=\linewidth]{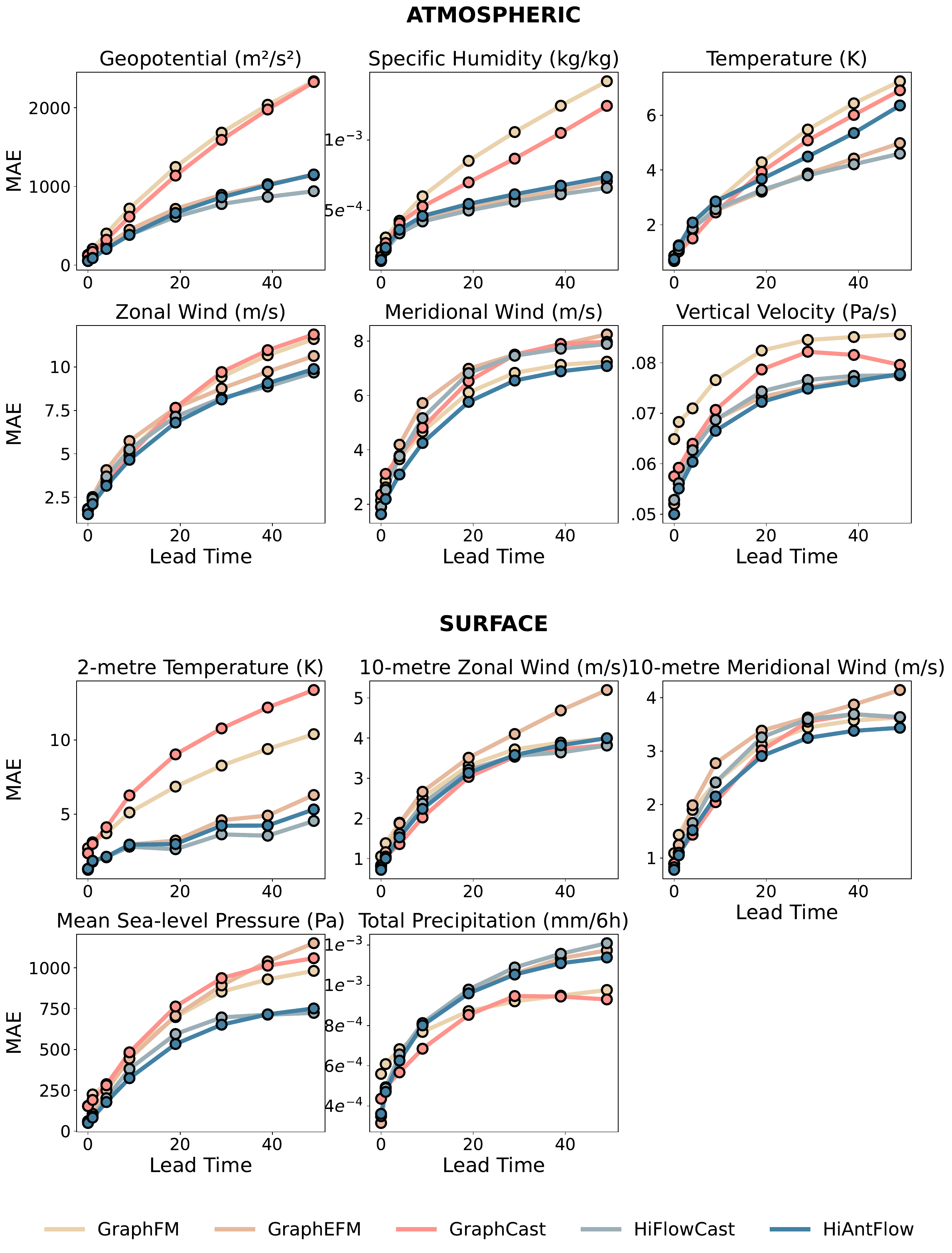}
    \caption{Average lead time MAE over starting timestamps and atmospheric levels for each atmospheric variable. Notably, HiFlowCast and HiAntFlow show strong performance over temperature, 2-metre temperature, geopotential, mean sea level pressure, and wind velocity variables.}
    \label{fig:mae_dis}
\end{figure*}

While aggregate metrics provide a high-level view of overall skill, disaggregated evaluations expose systematic biases across physical variables. We assess models independently over the five surface and six atmospheric variables listed in Table~\ref{tab:era5_vars}, with the latter averaged across the pressure levels in Table~\ref{tab:era5_pressure_levels}. Figure~\ref{fig:mae_dis} reports the mean absolute error (MAE) up to a 13-day lead time, corresponding to 50 sequential forecasts. The Flow models demonstrate marked improvements in geopotential, temperature, 2-metre temperature, and mean sea-level pressure. Moreover, at least one Flow variant achieves leading performance for Zonal and Meridional winds at both surface and atmospheric levels. Because all baselines embed solar radiation at a single resolution, the enhanced accuracy of the Flow models indicates stronger integration of the forcing term. By bridging latent and output representations, the Flow models more effectively propagate physical information across spatial scales, thereby improving multiscale process fidelity.

\subsection{Example Climate Projections}

We further our qualitative analysis from the main paper by examining intrinsic model behavior as it pertains to key variables. In particular, we plot examples of climate projections over Surface Temperature and Zonal and Meridional wind speeds. The latter exhibits intricate spatial patterns that are particularly challenging at long lead times. We sample at projections at $1$, $10$, and $13$-day lead times. 

\begin{figure*}[t]
    \centering
    \begin{subfigure}[t]{0.9\linewidth}
        \centering
        \includegraphics[width=\linewidth]{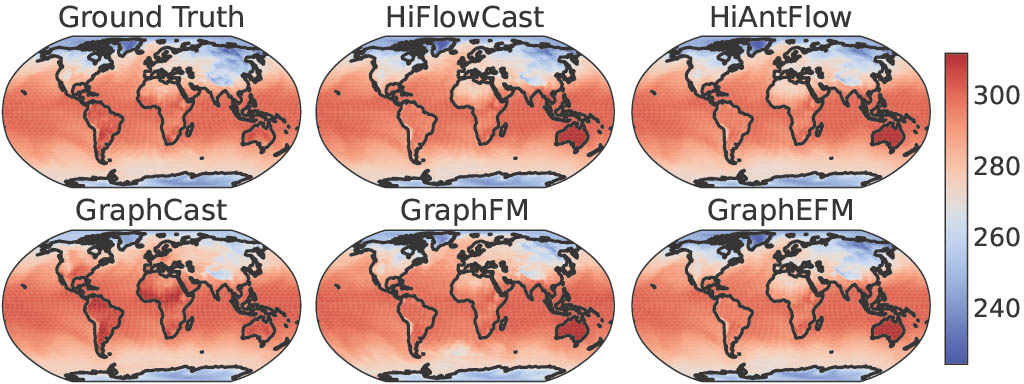}
        \caption{Lead time: 1 day}
    \end{subfigure}
    \vspace{20pt}

    \begin{subfigure}[t]{0.9\linewidth}
        \centering
        \includegraphics[width=\linewidth]{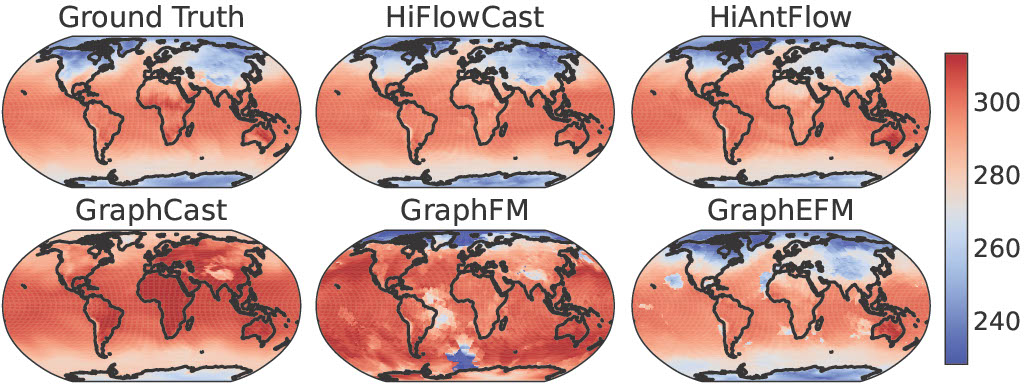}
        \caption{Lead time: 10 days}
    \end{subfigure}
    \vspace{20pt}

    \begin{subfigure}[t]{0.9\linewidth}
        \centering
        \includegraphics[width=\linewidth]{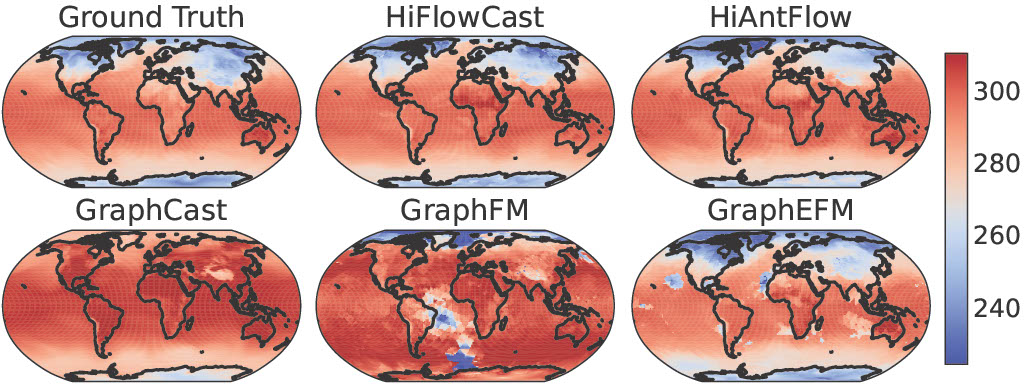}
        \caption{Lead time: 13 days}
    \end{subfigure}

    \caption{Projections of meridional wind speeds from all models at $1$, $5$, and $10$-days lead times at a pressure level of $500$ hPa. The sequence begins on the $8$th of January, 2020. Fidelity degrades as lead time increases.}
    \label{fig:surftemp_fig}
\end{figure*}

{\bf Surface Temperature.} Figure~\ref{fig:surftemp_fig} shows surface temperature forecasts up to a $13$-day lead time, initialized on 9 January 2020. GraphCast and GraphFM deteriorate rapidly, while GraphEFM loses physical plausibility at $10$ and $13$ days, with fields becoming noticeably less smooth. In contrast, the Flow models retain high fidelity to the ground truth throughout the forecast horizon. 

\begin{figure*}[t]
    \centering
    \begin{subfigure}[t]{0.9\linewidth}
        \centering
        \includegraphics[width=\linewidth]{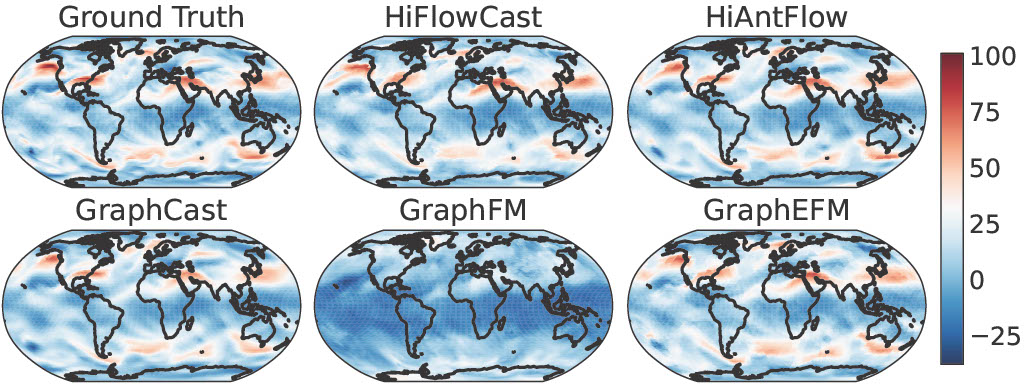}
        \caption{Lead time: 1 day}
    \end{subfigure}
    \vspace{20pt}

    \begin{subfigure}[t]{0.9\linewidth}
        \centering
        \includegraphics[width=\linewidth]{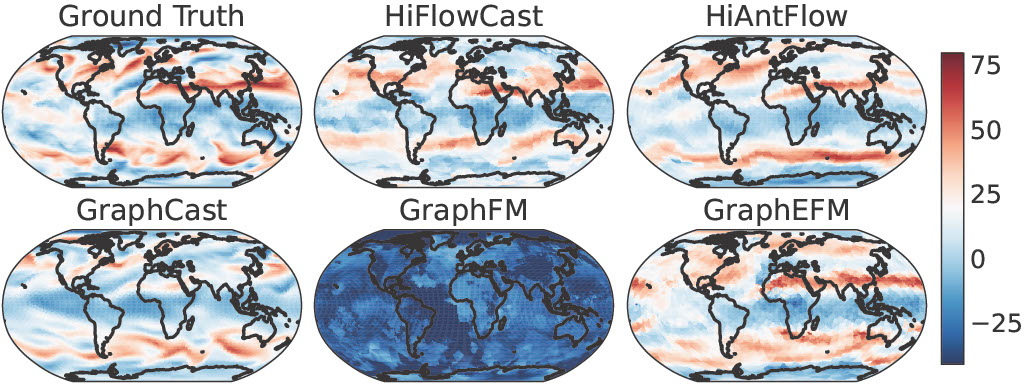}
        \caption{Lead time: 10 days}
    \end{subfigure}
    \vspace{20pt}

    \begin{subfigure}[t]{0.9\linewidth}
        \centering
        \includegraphics[width=\linewidth]{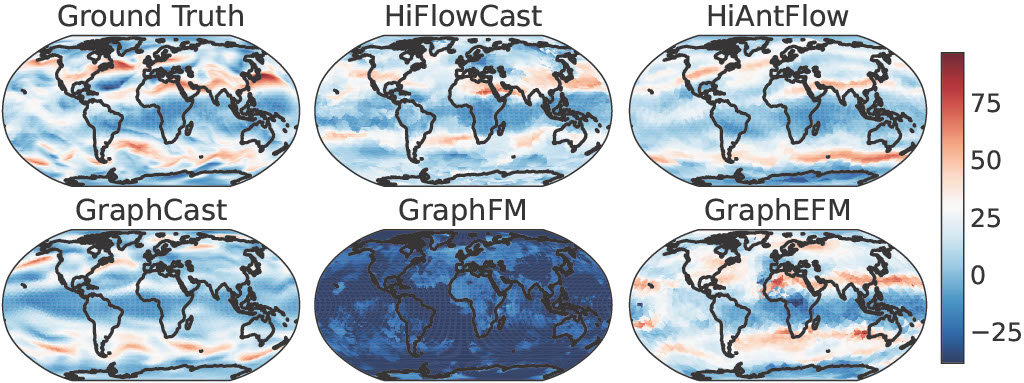}
        \caption{Lead time: 13 days}
    \end{subfigure}

    \caption{Projections of zonal wind speeds from all models at $1$, $10$, and $13$-day lead times at a pressure level of $500$ hPa. The sequence begins on the $8$th of January, 2020. Fidelity degrades as lead time increases.}
    \label{fig:uwindinstances}
\end{figure*}

{\bf Zonal wind velocity.} Figure~\ref{fig:uwindinstances} presents the evolution of model forecasts of the zonal wind velocity field at lead times of $1$, $10$, and $13$ days. HiFlowCast exhibits difficulty in preserving the fine-scale spatial structures characteristic of this field. By contrast, HiAntFlow demonstrates improved retention of these intricate patterns, capturing local variability with higher fidelity. GraphCast, meanwhile, provides consistently smooth and coherent projections, yielding a high-fidelity representation across lead times. 

\section{Code Availability.}

All code and model weights will be made publicly available via GitHub upon acceptance. Until then, we submit the code via the Code Ocean platform.

\end{appendices}

\end{document}